# A Benchmark for LLM's Understanding of Middle School and High School Science Topics

Noah L. Schroeder, Ph.D.[1*], Yessy Eka Ambarwati, M.A.E.[1], Yuji Zhang, Ph.D.[2], ChengXiang Zhai, Ph.D.[3]

[1]University of Florida; [2]College of Staten Island, City University of New York; [3]University of Illinois at Urbana-Champaign

* Correspondance should be addressed to Noah Schroeder: schroedern@ufl.edu

## Abstract

Large language models (LLMs) are increasingly integrated into educational settings, yet educators lack robust, standards-aligned tools to evaluate their effectiveness in K-12 science contexts. Existing benchmarks predominantly assess general language or advanced scientific reasoning, leaving a critical gap in understanding LLMs' performance on content directly relevant to secondary science curricula. To address this gap, we developed a comprehensive NGSS-aligned benchmark for both middle and high school science using a rigorous synthetic data pipeline, multi-judge validation, and item-level psychometric analysis. Nine open-weight LLMs were systematically evaluated using this benchmark, indicating that several smaller, locally deployable models achieved high accuracy across diverse science domains and question types. Our findings indicate that model size did not consistently predict performance, emphasizing the importance of intentional model selection for educational deployment. We then incorporated a human reviewer into the loop, reviewing the items generated by the LLMs for alignment with NGSS standards. The human review indicated that synthetically generated items were not in perfect alignment with the NGSS standards, indicating the benefits of human-in-the-loop item development, the need to explore the intersection of content and pedagogical knowledge, and the need to extend benchmarks to evaluate LLMs' capacity for interactive, evidence-based feedback in educational scenarios.

## 1. Introduction

Large Language models (LLMs) have rapidly emerged as a promising technology for supporting teaching and learning (Alhafni et al., 2024). Benefitting from a combination of semantic understanding and the ability to communicate in natural language, LLMs are quickly being integrated into tools that can support learners and teachers alike (Gan et al., 2023; Xu et al., 2023). LLMs can support a range of educational purposes, including generating instructional materials, providing feedback, tutoring, and personalized learning experiences (Kasneci et al., 2023; Tlili et al., 2023). This has led to growing incorporation of LLMs into educational technologies and instructional workflows. However, educators and researchers are faced with a non-trivial challenge: they must determine which LLM is most appropriate for their specific educational context, as LLM performance can vary substantially across domains, tasks, and reasoning demands (Pang et al., 2025; Wang et al., 2025).

One tool commonly used to evaluate and compare LLMs is a benchmark, a standardized set of tasks designed to assess performance across models using a shared evaluation framework (Reuel et al., 2024). Benchmarks play an important role in identifying model strengths and weaknesses, which in turn informs model selection decisions. Existing benchmarks largely focus on general language understanding (Hendrycks et al., 2020), coding and reasoning tasks, or advanced scientific problem solving (Wang et al., 2025). Recent benchmarking research has increasingly focused on scientific reasoning because performance on general language benchmarks may not adequately capture domain-specific knowledge and reasoning processes (Pang et al., 2025; Wang et al., 2025., Xu et al., 2025). Yet, most widely used benchmarks were not designed with K-12 education in mind, leaving educators with limited evidence regarding how well these systems perform on the science content students encounter in school. This presents challenges for educators and researchers alike, as we do not have a simple tool to help us decide which LLM to use in what educational situation.

From a science education perspective, LLMs should have deep knowledge aligned with established learning standards taught in schools (He et al., 2026). The Next Generation Science Standards (NGSS), developed from *A Framework for K-12 Science Education*, define science learning through performance expectations that integrate disciplinary core ideas, science and engineering practices, and crosscutting concepts (National Research Council, 2012; NGSS Lead States, 2013). Given its broad adoption in K-12 science education, the NGSS offer a well-grounded basis for evaluating LLM performance on standards-aligned science questions.

Accordingly, the purpose of this study is to develop and evaluate the first NGSS-aligned benchmark for middle school and high school science. By providing a standard-based evaluation framework, this study contributes empirical evidence regarding the capabilities and limitations of LLMs in science education and establishes a foundation for future educational AI benchmarks.

## 2. Literature Review

### *2.1 The State of LLMs in Education*

LLMs are often being deployed in both educational settings and educational research. A key distinction in the LLM landscape is between closed-weight and open-weight models. Closed-weight models are accessed only through an API, which is generally controlled by the company that built them. Alternatively, open-weight models are those that have their model weights released publicly, so they can be deployed and modified by anyone with the necessary infrastructure. This difference matters in practice for educators: among other benefits (e.g., fine-tuning and other post-training opportunities), open-weight models can be hosted on local servers, enabling educators and researchers to retain control over student data (Overton et al., 2025; Schroeder, 2026). However, the practicality of open weight deployment depends on model size and computational requirement (Zhu et al., 2024). This is essential in education, where data governance and control over deployment are central to responsible AI use (Schroeder et al., 2026).

Concerns about student data privacy with AI tools echo earlier debates in educational data mining, where researchers stressed that student data (grades, demographics, or similar records) are sensitive and need to be protected (Vaccaro et al., 2024). Open-weight models can help address this because they can be run locally down to the device level, meaning each student could work directly with their own LLM and the data never leaves their computer. Such an approach reduces reliance on outside services, giving individuals full control over their data (Schroeder at al., 2026). In addition to privacy benefits, local deployment of open-weight models may lower recurring API costs and facilitate scalable deployment across classrooms and schools (Glenn & Samuel, 2026)

When discussing LLMs, one must note that each model may differ in architecture, parameter size, training procedures, and performance characteristics, and these differences affect model capabilities, efficiency, reliability, and suitability in different classroom scenarios and contexts (Shao et al., 2024; Wang et al., 2026). For example, one model may be very knowledgeable about scientific concepts, whereas another may be more skilled at writing. These inherent differences are due to the training data and pipelines used to create the models, but practically mean that educators and researchers should require evidence about how each model performs on relevant educational tasks before recommending them for classroom use. Benchmarking provides a systematic way to generate that evidence, comparing models under a shared framework to support better-informed model selection (Reuel et al., 2024).

### *2.2 Benchmarking as Educational Measurement*

We conceptualize benchmarking as a conceptual parallel to educational measurement. Measurement instruments are designed to generate evidence that supports inferences about performance with a defined domain (Kane, 2013; Messick, 1989). Similar to how assessments are used to evaluate student learning, benchmarks operationalize a target knowledge domain through a structured collection of tasks and generates performance evidence that supports interpretive claims about ability or competence (Kane, 2013; Mislevy et al., 2003). Practically speaking, LLM benchmarks often consist of question and answer pairs that function as a ground truth, from which any individual model is tested against to see how accurate their answers are in comparison to the ground truth dataset.

Consequently, benchmark scores are meaningful only to the extent that benchmark tasks adequately represent the intended domain(s), akin to the construct of validity in the educational measurement space (Freiesleben & Zezulka, 2025). For this work, we argue that our benchmark scores reflect observed *performance* rather than underlying *competence*. Patterns of performance across content areas may reveal meaningful distinctions that are obscured by a single aggregate score, so we report fine-grained and summary data for each model tested.

Arguments around various forms of validity often result in conversations around item-level quality metrics and alignment with the construct being examined. We leverage these conventions in our view of educational benchmarks for LLMs.

*2.3 Benchmark Quality and Item-Level Evaluation*

Educational measurement provides useful approaches for evaluating benchmark quality. Two commonly used psychometric indicators are item discrimination and item difficulty. Item discrimination refers to the extent to which an item differentiates between high-performing and low-performing respondents, whereas item difficulty reflects the proportion of respondents who answer an item correctly (Ebel & Frisbie, 1991). By examining these metrics, we can help ensure that the benchmark is not trivial for LLMs to complete (Schroeder et al., 2026), as that would not allow for researchers to identify models that perform better or worse in different areas of the NGSS.

Researchers have emphasized the importance of avoiding benchmarks that are either too easy, too difficult, or insufficiently discriminating, as such benchmarks provide limited information about differences among models (Reuel et al., 2024; Schroeder et al., 2026). Similarly, Lelievre et al. (2025) highlight concerns regarding benchmark saturation, where increasingly capable models reach near-ceiling performance, limiting the benchmark's ability to differentiate between systems. Together, evidence suggests that it is critical that item-level metrics are examined in any benchmark creation task to ensure one can make reasonable inferences about model performance on the relevant task.

*2.4 Existing Educational Benchmarks for LLMs*

Examining the literature shows that few educational benchmarks for LLMs have been published to date. However, those that do exist can generally be categorized as either measuring content knowledge, or pedagogical knowledge and abilities.

Measuring content knowledge has been one approach used to create educational benchmarks. For example, Ali et al. (2024) examined how well LLMs scored on various concept inventories in computer science. Other research teams have taken a broader approach, measuring multiple content areas. E-EVAL is a good example of this, measuring a variety of subjects across Chinese K-12 education (Hou et al., 2024). However, research in this area is relatively sparse compared to benchmarks around pedagogical knowledge and teaching abilities.

The benchmarks around teaching-focused tasks have taken a variety of perspectives on what is important to measure. For example, Lelievre et al. (2025) developed a benchmark based on the pedagogical tests given to teachers in Chile, while Xu et al. (2026) measured how well LLMs can adapt to various scenarios, provide accurate results, and follow pedagogical principles. The literature also shows a number of other approaches to benchmarking as well, such as measuring LLMs question generation ability (Chen et al., 2024; Jaldi et al., 2026) and various teaching-related tasks (Wei et al., 2025).

*2.5 Synthetic Data Generation and LLM-as-a-Judge*

Beyond serving as subjects of evaluation, LLMs are increasingly being used as a tool within benchmark construction flows (Long et al., 2024). One emerging application is synthetic data generation, in which LLMs generate questions, responses, labels, or other forms of structured data for training and evaluation purposes (Long et al., 2024; Chen et al., 2024). Synthetic data can support dataset expansion, reduce content development time, and generate benchmark items in domains where expert-authored content is limited. Recent work has demonstrated the value of synthetic data for data augmentation (Wang et al., 2024), benchmark construction (Chen et al., 2024; Xu et al., 2024), and educational assessment (Jaldi et al., 2026), although concerns remain regarding data quality, validity, and potential propagation of biases present in the underlying models (Chen et al., 2024).

Another emerging application is the use of LLMs as evaluators, commonly referred to as LLM-as-a-Judge. In this approach, LLMs assess outputs produced by humans or other language models and have been used for tasks such as response quality evaluation, automated assessment, benchmark development, and educational evaluation (Gu et al., 2025; Het et al., 2026). The growing interest in LLM judges is driven in part by the limitations of traditional human evaluation, which can be costly, time-consuming, and difficult to scale (Gu et al., 2025). Additionally, LLM judges can generate explanations, rationales, and feedback to support their evaluations (Gu et al., 2025; He et al., 2026). Therefore, LLM judges offer advantages in scalability, consistency, and cost-compared with fully human evaluation processes and have shown substantial agreement with human raters across a range of evaluation tasks (Gu et al., 2025; Zheng et al., 2023). However, concerns remain regarding evaluator bias, reliability, and agreement with human judgements, making human validation an important consideration when deploying LLM-based evaluation systems (Bavaresco et al., 2025; Chen et al., 2024; He et al., 2026).

In this study, we leverage LLMs for both synthetic data generation as well as acting as judges. Given the limitations they have in both regards, we have incorporated a human in the loop process, as described in Part III of the manuscript.

*2.6 The Present Study*

Together, we can see that the field has some existing benchmarks, but they have largely been focused on pedagogical abilities rather than content knowledge specifically. Moreover, we see few examples of benchmarks aligned with specific educational content standards. In this study, we address this gap in the literature by creating benchmarks aligned with the widely-known NGSS, examine item-level performance across the benchmark, and examine model performance across a variety of question types. Specifically, this study consists of three parts.

In Part I, we develop a ground-truth dataset framed around the NGSS. This consisted of a comprehensive synthetic data production pipeline with a multiple-judge framework, as well as a graphic user interface (GUI) for navigating the pipeline and results.

In Part II, we benchmark 9 open-weight LLMs to gain an understanding of their performance in secondary science. Specifically, we address the following research questions:

RQ1: How well do popular open-weight LLMs understand secondary science topics across various item types?

RQ2: How well did the individual items in the benchmark function?

In Part III, we address human validation of items. Specifically, we address the following research question:

RQ3: To what extent does a former K-12 science teacher feel the items align with the NGSS standards?

## 3. Part I

Before we can benchmark LLMs, we first had to develop a dataset to use as the benchmark. For this study, we focused on the NGSS for secondary science.

### 3.1 Methods

The same methods were used for creating the middle school and high school standards-aligned dataset, unless noted otherwise.

#### *3.1.1 Dataset Generation*

To construct a robust ground-truth dataset, we utilized Sonnet 4.6 Thinking to generate question–answer pairs aligned with the 12 middle school science areas defined by the NGSS. Without any prior hypothesis, we targeted a balanced distribution across three cognitive levels: recall (34 items), conceptual understanding (33 items), and applied reasoning (33 items). Similarly, we created question-answer pairs aligned with the 12 high school science areas in the NGSS, and distributed questions across recall (33 questions), conceptual understanding (33 items), and applied reasoning (34 items).

#### *3.1.2. Item Quality Requirements*

To help ensure high data quality, each candidate item underwent a three-stage automated filtering process. First, near-duplicate entries were removed using the rapidfuzz token-set ratio with a threshold of ≥ 88. Second, we enforced a middle school readability constraint using the Flesch–Kincaid grade level (targeting a range of 5.0–9.0) for the middle school dataset, and a reading level constraint of 5.0-13.0 for the high school dataset. Finally, an item-level LLM-based appropriateness check was conducted.

#### *3.1.3 Item Validation*

The resulting items were then subjected to a formal validation process. A three-judge panel, consisting of Nemotron 3 Super 120b a12, Gemini Flash 3, and OSS 120b, independently scored each question–answer pair. They addressed the accuracy of the question as well as if the question-answer pair met the standard it was meant to align with. To maintain the integrity of the benchmarks, any questions that elicited disagreement among the judges or were unanimously rejected were automatically removed from the set.

#### *3.1.4 Code Generation*

All code used to power this study was developed with the assistance of Claude LLMs via Claude Code within Visual Studio Code. Primarily Opus 4.8/5 was used, but Sonnet 4.5/5 was used for smaller, more well-defined coding tasks.

*3.1.5 Data and Code Availability*

All data and code to support this work are available on github: Middle School: https://github.com/InviteInstitute/MS-NGSS-Aligned-Benchmark High School: https://github.com/InviteInstitute/HS-NGSS-Aligned-Benchmark

## 3.2 Results

*3.2.1 Dataset Generation*

Our pipeline resulted in the rejection and subsequent regeneration of 4,433 candidate questions when developing the middle school dataset. The high school dataset creation resulted in the rejection and regeneration of 1,770 items.

*3.2.2 Item Validation*

The judge models disagreed on 45 and 44 items on the middle school and high school datasets, respectively. These items were subsequently dropped from the benchmark. These disagreements and the details of each are presented in tabular form in the respective github repositories.

*3.2.3 Final Dataset*

Our final dataset consisted of 1,078 items for middle school benchmark, and 1,150 items for the high school benchmark. The distribution of questions across standards is in Table 1, while the distribution of questions by area, standard, and question type is provided in the github repositories.

Table 1. Distribution of questions per area of the NGSS.

Middle School

| Area | Recall | Conceptual | Applied | Total |
|---|---|---|---|---|
| MS-PS1 | 23 | 33 | 31 | 87 |
| MS-PS2 | 29 | 31 | 31 | 91 |
| MS-PS3 | 22 | 30 | 32 | 84 |
| MS-PS4 | 29 | 28 | 31 | 88 |
| MS-LS1 | 29 | 31 | 29 | 89 |
| MS-LS2 | 31 | 32 | 33 | 96 |
| MS-LS3 | 26 | 29 | 30 | 85 |
| MS-LS4 | 27 | 33 | 29 | 89 |
| MS-ESS1 | 31 | 33 | 32 | 96 |
| MS-ESS2 | 31 | 31 | 32 | 94 |
| MS-ESS3 | 26 | 33 | 32 | 91 |
| MS-ETS1 | 25 | 32 | 31 | 88 |
| Total | 329 | 376 | 373 | 1078 |

High School

| Area | Recall | Conceptual | Applied | Total |
|---|---|---|---|---|
| HS-PS1 | 33 | 33 | 33 | 99 |
| HS-PS2 | 32 | 33 | 34 | 99 |
| HS-PS3 | 31 | 30 | 32 | 93 |
| HS-PS4 | 32 | 32 | 34 | 98 |
| HS-LS1 | 32 | 33 | 33 | 98 |
| HS-LS2 | 31 | 29 | 28 | 88 |
| HS-LS3 | 32 | 31 | 31 | 94 |
| HS-LS4 | 31 | 32 | 32 | 95 |
| HS-ESS1 | 31 | 32 | 32 | 95 |
| HS-ESS2 | 31 | 33 | 32 | 96 |
| HS-ESS3 | 32 | 32 | 34 | 98 |
| HS-ETS1 | 31 | 32 | 34 | 97 |
| Total | 379 | 382 | 389 | 1150 |

### 3.3 Data Infrastructure Created

We set out to create a benchmark aligned with the NGSS for secondary science. Using a four-tier data quality gate system, we removed duplicate items, ensured each item had an appropriate reading level, used appropriate question wording, and that three different LLM's agreed with not only the answer to each question in the ground-truth dataset, but also ensured it met the standard it was designed to align with. This pipeline resulted in a comprehensive ground-truth dataset to make up our benchmarks. The dataset enabled us to study how well LLMs scored on this benchmark and how each item performed.

## 4. Part II

With a ground-truth dataset in hand, our next step was to evaluate a variety of open-weight LLMs to gain insights as to their performance in middle school science. We emphasize open-weight LLMs, and primarily those under 100b parameters, as they are available for free and are reasonably-sized for local deployment in educational contexts. This allows an individual, school, or district to maintain strict control over the data pipeline which is necessary to ensure the protection of student data.

## 4.1. Methods

### *4.1.1 Dataset*

We used the datasets created in Part I.

### *4.1.2 LLMs Evaluated*

The evaluation procedure involved nine representative candidate models released by six different industry labs, each tasked with answering every question in the dataset three times (designated as runs 1, 2, and 3). The models represented a variety of open-weight model families and models of various sizes and architectures. Specifically, we benchmarked (ordered by model size): Llama 3.1 8b, Granite 3.3 8b Instruct, OSS-20b, Mistral Small 3.1, Gemma 3 27b, Gemma 4 31b, Nemotron 3 Nano 30b A3, Llama 3.3 70b, and OSS-120b.

### *4.1.3 Prompting Method*

We employed a standardized zero-shot protocol, utilizing a shared middle school or high school prompt, respectively, for all models to ensure consistency. No per-model tuning or specialized prompting was used between models or model families.

### *4.1.4 Evaluation Procedures*

Responses were evaluated by a single judge (Nemotron 3 Super 120b A12) using a binary correct/incorrect metric for recall items, and a three-point rubric (0, 0.5, or 1) for conceptual and applied items. Any responses that were empty, errored, or constituted a refusal to answer were assigned a score of 0 and are reported separately as the model's refusal rate. We also report and distinguish between macro accuracy, which weights each NGSS science area equally, and micro accuracy, which weights each individual question equally.

### *4.1.5 Single-shot and Ensemble Methods*

We utilized two primary scoring conventions to characterize model performance, single-shot and ensemble methods. Single-shot scores represent accuracy on the first run only. The ensemble score represents accuracy determined by a majority vote across all three runs, serving as a measure of the model's self-consistency. All accuracies are reported as proportions between 0 and 1.

### *4.1.6 Item-level Performance*

We evaluated the item difficulty and item discrimination for every item. We also classified items as trivial for the models (where 95% or more of answers were correct) or broken (where 5% or less of the models answered correctly).

*4.1.7 Data and Code Availability*

All data and code to support this work are available on github in the same repositories linked in Part I.

## 4.2 Results

Extensive reports of primarily tabular results are provided for the middle school and high school level benchmarks independently in the github repository. We report only the most pressing results here.

### 4.2.1 RQ1: How well do popular open-weight LLMs understand secondary science topics across various item types?

The overall results of the benchmarking are presented in Table 2. We observe that OSS 120b performed best across both benchmarks, and most of the models tested here scored above 90% on both benchmarks. Using majority-vote ensemble methods did not cause drastic improvements in scores on either benchmark, with increases from 1% to 5% being observed. We did not observe any model score degradation from using ensemble methods.

**Table 2**

Leaderboard of model performance across the middle school and high school benchmarks.

**Middle School**

| Rank | Model | One-Shot | Ensemble | Ensemble Gain | Mean Response Reading Level | Mean Total Tokens |
|---|---|---|---|---|---|---|
| 1 | OSS 120b | 0.98 | 0.99 | 0.01 | 8.16 | 611.20 |
| 2 | Nemotron 3 Nano 30ba3 | 0.97 | 0.99 | 0.02 | 7.83 | 433.70 |
| 3 | Gemma 4 31b | 0.98 | 0.99 | 0.01 | 7.42 | 315.50 |
| 4 | OSS 20b | 0.94 | 0.97 | 0.04 | 7.65 | 556.10 |
| 5 | Mistral Small 3.1 | 0.91 | 0.95 | 0.05 | 7.48 | 314.50 |
| 6 | Llama 3.3 70b | 0.93 | 0.95 | 0.02 | 7.49 | 282.60 |
| 7 | Gemma 3 27b | 0.92 | 0.93 | 0.01 | 6.46 | 273.00 |
| 8 | Granite 3.3 8b Instruct | 0.83 | 0.88 | 0.05 | 9.35 | 308.60 |
| 9 | Llama 3.1 8b | 0.80 | 0.84 | 0.04 | 7.56 | 335.00 |

**High School**

| Rank | Model | One-Shot | Ensemble | Ensemble Gain | Mean Response Reading Level | Mean Total Tokens |
|---|---|---|---|---|---|---|
| 1 | OSS 120b | 0.98 | 1.00 | 0.01 | 11.28 | 895.70 |
| 2 | Gemma 4 31b | 0.98 | 0.99 | 0.01 | 11.41 | 492.30 |
| 3 | Nemotron 3 Nano 30ba3 | 0.97 | 0.98 | 0.01 | 12.12 | 539.00 |
| 4 | OSS 20b | 0.96 | 0.98 | 0.02 | 11.07 | 723.60 |
| 5 | Gemma 3 27b | 0.96 | 0.97 | 0.01 | 11.62 | 374.10 |
| 6 | Mistral Small 3.1 | 0.91 | 0.97 | 0.05 | 11.66 | 411.10 |
| 7 | Llama 3.3 70b | 0.93 | 0.96 | 0.02 | 12.76 | 365.50 |
| 8 | Granite 3.3 8b Instruct | 0.81 | 0.86 | 0.05 | 12.71 | 435.30 |
| 9 | Llama 3.1 8b | 0.78 | 0.83 | 0.05 | 11.86 | 426.30 |

We next examined model performance by question type (Table 3). We found the most accurate models on the benchmark overall tended to score consistently well across item types.

An important consideration is the computational efficiency to accuracy tradeoff, considering both model size and average total tokens used. There are really two competitors in this space, OSS-20b, with a one-shot accuracy of 94% (middle school) to 96% (high school), and Nemotron 3 Nano 30b a3 which had one-shot accuracy of 97% on both benchmarks. The differences between the models' accuracy were not statistically significant using pairwise McNemar tests with Holm-Bonferroni corrections at the high school level, but they were at the middle school level, with Nemotron scoring significantly better. In both benchmarks, the Nemotron model used less tokens on average.

**Table 3**

Accuracy by question type for both single-shot and ensemble methods.

**Middle School**

| Model | Recall (single) | Recall (ensemble) | Conceptual (single) | Conceptual (ensemble) | Applied (single) | Applied (ensemble) |
|---|---|---|---|---|---|---|
| OSS 120b | 0.99 | 0.99 | 0.99 | 1.00 | 0.98 | 0.99 |
| Nemotron 3 Nano 30ba3 | 0.97 | 0.99 | 0.97 | 0.99 | 0.96 | 0.98 |
| Gemma 4 31b | 0.97 | 0.98 | 0.98 | 0.99 | 0.98 | 0.99 |
| OSS 20b | 0.94 | 0.98 | 0.93 | 0.97 | 0.94 | 0.97 |
| Mistral Small 3.1 | 0.90 | 0.96 | 0.90 | 0.94 | 0.91 | 0.95 |
| Llama 3.3 70b | 0.94 | 0.96 | 0.91 | 0.93 | 0.94 | 0.96 |
| Gemma 3 27b | 0.93 | 0.92 | 0.92 | 0.95 | 0.91 | 0.92 |
| Granite 3.3 Instruct | 0.88 | 0.92 | 0.82 | 0.88 | 0.80 | 0.84 |
| Llama 3.1 8b | 0.85 | 0.89 | 0.77 | 0.80 | 0.78 | 0.82 |

**High School**

| Model | Recall (single) | Recall (ens) | Conceptual (single) | Conceptual (ens) | Applied (single) | Applied (ens) |
|---|---|---|---|---|---|---|
| OSS 120b | 0.99 | 1.00 | 0.99 | 1.00 | 0.97 | 0.99 |
| Gemma 4 31b | 0.98 | 0.98 | 0.98 | 1.00 | 0.99 | 0.99 |
| Nemotron 3 Nano 30ba3 | 0.97 | 0.97 | 0.97 | 0.99 | 0.97 | 0.99 |
| OSS 20b | 0.96 | 0.98 | 0.96 | 0.99 | 0.96 | 0.98 |
| Gemma 3 27b | 0.98 | 0.98 | 0.97 | 0.98 | 0.94 | 0.96 |
| Mistral Small 3.1 | 0.91 | 0.97 | 0.93 | 0.97 | 0.90 | 0.96 |
| Llama 3.3 70b | 0.96 | 0.98 | 0.91 | 0.93 | 0.93 | 0.96 |
| Granite 3.3 Instruct | 0.87 | 0.93 | 0.80 | 0.84 | 0.77 | 0.82 |
| Llama 3.1 8b | 0.83 | 0.89 | 0.75 | 0.79 | 0.75 | 0.80 |

Lastly, we examined the ensemble scores for each model across each NGSS standard (Table 4). The most important takeaway from these results is that models that score well on the benchmark overall tended to score well across all areas of the benchmark. Meanwhile, models that scored worse across the benchmarks (e.g., Llama 3.1 8b, Granite 3.3 Instruct) had notable variation across content areas.

**Table 4**

Ensemble scores by model across each content standard area.

**Middle School**

| Model | MS-PS1 | MS-PS2 | MS-PS3 | MS-PS4 | MS-LS1 | MS-LS2 | MS-LS3 | MS-LS4 | MS-ESS1 | MS-ESS2 | MS-ESS3 | MS-ETS1 |
|---|---|---|---|---|---|---|---|---|---|---|---|---|
| OSS 120b | 1.00 | 0.97 | 1.00 | 1.00 | 0.99 | 0.98 | 1.00 | 1.00 | 1.00 | 1.00 | 0.99 | 1.00 |
| Nemetron 3 Nano 30ba3 | 0.98 | 0.98 | 1.00 | 1.00 | 0.98 | 0.99 | 1.00 | 1.00 | 0.99 | 0.99 | 0.98 | 0.99 |
| Gemma 4 31b | 0.98 | 1.00 | 1.00 | 0.99 | 0.98 | 0.97 | 1.00 | 1.00 | 0.98 | 0.98 | 0.99 | 0.99 |
| OSS 20b | 0.99 | 0.95 | 0.99 | 0.97 | 0.98 | 0.99 | 0.99 | 0.93 | 0.98 | 0.94 | 0.99 | 1.00 |
| Mistral Small 3.1 | 0.94 | 0.92 | 0.93 | 0.93 | 0.93 | 0.98 | 0.98 | 0.96 | 0.93 | 0.95 | 0.99 | 0.99 |
| Llama 3.370b | 0.90 | 0.91 | 0.92 | 0.96 | 0.96 | 0.96 | 0.99 | 0.97 | 0.96 | 0.94 | 0.98 | 0.99 |
| Gemma 327b | 0.87 | 0.90 | 0.99 | 0.98 | 0.99 | 0.96 | 0.93 | 0.96 | 0.94 | 0.90 | 0.86 | 0.91 |
| Granite 3.3 Instruct | 0.77 | 0.80 | 0.83 | 0.91 | 0.91 | 0.94 | 0.87 | 0.88 | 0.79 | 0.90 | 0.97 | 0.96 |
| Llama 3.18b | 0.72 | 0.80 | 0.82 | 0.82 | 0.85 | 0.90 | 0.89 | 0.82 | 0.70 | 0.79 | 0.93 | 0.99 |

**High School**

| Model | HS-PS1 | HS-PS2 | HS-PS3 | HS-PS4 | HS-LS1 | HS-LS2 | HS-LS3 | HS-LS4 | HS-ESS1 | HS-ESS2 | HS-ESS3 | HS-ETS1 |
|---|---|---|---|---|---|---|---|---|---|---|---|---|
| OSS 120b | 1.00 | 1.00 | 1.00 | 1.00 | 1.00 | 1.00 | 1.00 | 1.00 | 1.00 | 0.98 | 0.98 | 0.99 |
| Gemma 4 31b | 0.98 | 1.00 | 0.98 | 1.00 | 0.99 | 1.00 | 1.00 | 1.00 | 0.98 | 0.96 | 1.00 | 1.00 |
| Nemotron 3 Nano 30ba3 | 0.99 | 1.00 | 0.98 | 0.97 | 0.98 | 0.98 | 0.99 | 0.97 | 0.97 | 0.99 | 0.99 | 0.99 |
| OSS 20b | 0.98 | 1.00 | 0.97 | 1.00 | 0.99 | 0.98 | 1.00 | 0.95 | 1.00 | 0.96 | 0.97 | 1.00 |
| Gemma 3 27b | 0.99 | 0.99 | 0.94 | 0.98 | 0.96 | 0.97 | 0.99 | 0.97 | 0.95 | 0.98 | 0.99 | 0.99 |
| Mistral Small 3.1 | 0.96 | 0.99 | 0.93 | 0.98 | 0.96 | 0.96 | 1.00 | 0.96 | 0.95 | 0.94 | 0.98 | 0.99 |
| Llama 3.3 70b | 1.00 | 0.97 | 0.95 | 0.99 | 0.93 | 0.92 | 0.94 | 0.97 | 0.92 | 0.92 | 0.98 | 0.98 |
| Granite 3.3 Instruct | 0.73 | 0.89 | 0.84 | 0.93 | 0.83 | 0.92 | 0.79 | 0.96 | 0.75 | 0.80 | 0.95 | 0.97 |
| Llama 3.1 8b | 0.73 | 0.84 | 0.90 | 0.91 | 0.83 | 0.88 | 0.77 | 0.83 | 0.66 | 0.68 | 0.96 | 0.94 |

#### 4.2.2 RQ2: How well did the individual items in the benchmark function?

We next examined mean item-level metrics by standard and question type (Table 5). As shown, generally speaking, we observed high mean difficulty and low item discrimination. There was only one standard and question type (MS-LS2, Recall) with a single broken question at the middle school level, but we saw a range of trivial questions depending on the standard and question type. At times, the number of trivial questions exceeded 90%.

**Table 5**

Mean item-level by standard and question type.

**Middle School**

| Area | Question Type | *n* Items | Mean Difficulty | Mean Item Discrimination | % Trivial | % Broken |
|---|---|---|---|---|---|---|
| MS-PS1 | applied | 31 | 0.90 | 0.16 | 0.58 | 0.00 |
| MS-PS1 | conceptual | 33 | 0.88 | 0.21 | 0.46 | 0.00 |
| MS-PS1 | recall | 23 | 0.95 | 0.11 | 0.70 | 0.00 |
| MS-PS2 | applied | 31 | 0.86 | 0.20 | 0.48 | 0.00 |
| MS-PS2 | conceptual | 31 | 0.92 | 0.16 | 0.55 | 0.00 |
| MS-PS2 | recall | 29 | 0.97 | 0.07 | 0.79 | 0.00 |
| MS-PS3 | applied | 32 | 0.95 | 0.12 | 0.63 | 0.00 |
| MS-PS3 | conceptual | 30 | 0.93 | 0.14 | 0.63 | 0.00 |
| MS-PS3 | recall | 22 | 0.95 | 0.11 | 0.68 | 0.00 |
| MS-PS4 | applied | 31 | 0.94 | 0.14 | 0.61 | 0.00 |
| MS-PS4 | conceptual | 28 | 0.93 | 0.17 | 0.50 | 0.00 |
| MS-PS4 | recall | 29 | 0.99 | 0.04 | 0.90 | 0.00 |
| MS-LS1 | applied | 29 | 0.95 | 0.07 | 0.83 | 0.00 |
| MS-LS1 | conceptual | 31 | 0.96 | 0.10 | 0.71 | 0.00 |
| MS-LS1 | recall | 29 | 0.94 | 0.06 | 0.83 | 0.00 |
| MS-LS2 | applied | 33 | 0.98 | 0.03 | 0.91 | 0.00 |
| MS-LS2 | conceptual | 32 | 0.95 | 0.10 | 0.72 | 0.00 |
| MS-LS2 | recall | 31 | 0.95 | 0.04 | 0.84 | 0.03 |
| MS-LS3 | applied | 30 | 0.96 | 0.09 | 0.73 | 0.00 |
| MS-LS3 | conceptual | 29 | 0.95 | 0.11 | 0.69 | 0.00 |
| MS-LS3 | recall | 26 | 0.97 | 0.08 | 0.77 | 0.00 |
| MS-LS4 | applied | 29 | 0.97 | 0.08 | 0.76 | 0.00 |
| MS-LS4 | conceptual | 33 | 0.94 | 0.14 | 0.61 | 0.00 |
| MS-LS4 | recall | 27 | 0.92 | 0.14 | 0.67 | 0.00 |
| MS-ESS1 | applied | 32 | 0.89 | 0.24 | 0.34 | 0.00 |
| MS-ESS1 | conceptual | 33 | 0.91 | 0.17 | 0.55 | 0.00 |
| MS-ESS1 | recall | 31 | 0.96 | 0.11 | 0.68 | 0.00 |
| MS-ESS2 | applied | 32 | 0.92 | 0.16 | 0.56 | 0.00 |
| MS-ESS2 | conceptual | 31 | 0.95 | 0.11 | 0.71 | 0.00 |
| MS-ESS2 | recall | 31 | 0.93 | 0.15 | 0.58 | 0.00 |
| MS-ESS3 | applied | 32 | 0.94 | 0.09 | 0.75 | 0.00 |
| MS-ESS3 | conceptual | 33 | 0.99 | 0.04 | 0.88 | 0.00 |
| MS-ESS3 | recall | 26 | 0.96 | 0.11 | 0.69 | 0.00 |
| MS-ETS1 | applied | 31 | 0.97 | 0.08 | 0.77 | 0.00 |
| MS-ETS1 | conceptual | 32 | 0.98 | 0.06 | 0.81 | 0.00 |
| MS-ETS1 | recall | 25 | 1.00 | 0.01 | 0.96 | 0.00 |
| **High School** | | | | | | |
| **Area** | **Question Type** | ***n* Items** | **Mean Difficulty** | **Mean Item Discrimination** | **% Trivial** | **% Broken** |
| HS-PS1 | applied | 33 | 0.93 | 0.17 | 0.52 | 0.00 |
| HS-PS1 | conceptual | 33 | 0.93 | 0.17 | 0.52 | 0.00 |
| HS-PS1 | recall | 33 | 0.93 | 0.17 | 0.52 | 0.00 |
| HS-PS2 | applied | 34 | 0.97 | 0.06 | 0.82 | 0.00 |
| HS-PS2 | conceptual | 33 | 0.95 | 0.12 | 0.64 | 0.00 |
| HS-PS2 | recall | 32 | 0.97 | 0.08 | 0.78 | 0.00 |
| HS-PS3 | applied | 32 | 0.93 | 0.13 | 0.66 | 0.00 |
| HS-PS3 | conceptual | 30 | 0.91 | 0.12 | 0.70 | 0.00 |
| HS-PS3 | recall | 31 | 0.99 | 0.03 | 0.90 | 0.00 |
| HS-PS4 | applied | 34 | 0.97 | 0.06 | 0.82 | 0.00 |
| HS-PS4 | conceptual | 32 | 0.95 | 0.12 | 0.63 | 0.00 |
| HS-PS4 | recall | 32 | 1.00 | 0.01 | 0.97 | 0.00 |
| HS-LS1 | applied | 33 | 0.94 | 0.15 | 0.58 | 0.00 |
| HS-LS1 | conceptual | 33 | 0.92 | 0.12 | 0.70 | 0.00 |
| HS-LS1 | recall | 32 | 0.96 | 0.08 | 0.75 | 0.00 |
| HS-LS2 | applied | 28 | 0.93 | 0.11 | 0.71 | 0.00 |
| HS-LS2 | conceptual | 29 | 0.97 | 0.07 | 0.79 | 0.00 |
| HS-LS2 | recall | 31 | 0.96 | 0.07 | 0.81 | 0.00 |
| HS-LS3 | applied | 31 | 0.91 | 0.21 | 0.39 | 0.00 |

| HS-LS3 | conceptual | 31 | 0.94 | 0.15 | 0.58 | 0.00 |
|---|---|---|---|---|---|---|
| HS-LS3 | recall | 32 | 0.97 | 0.07 | 0.81 | 0.00 |
| HS-LS4 | applied | 32 | 0.98 | 0.05 | 0.88 | 0.00 |
| HS-LS4 | conceptual | 32 | 0.97 | 0.08 | 0.75 | 0.00 |
| HS-LS4 | recall | 31 | 0.92 | 0.12 | 0.71 | 0.00 |
| HS-ESS1 | applied | 32 | 0.87 | 0.23 | 0.41 | 0.00 |
| HS-ESS1 | conceptual | 32 | 0.89 | 0.23 | 0.38 | 0.00 |
| HS-ESS1 | recall | 31 | 0.96 | 0.06 | 0.84 | 0.00 |
| HS-ESS2 | applied | 32 | 0.91 | 0.15 | 0.56 | 0.00 |
| HS-ESS2 | conceptual | 33 | 0.89 | 0.20 | 0.49 | 0.00 |
| HS-ESS2 | recall | 31 | 0.93 | 0.13 | 0.65 | 0.00 |
| HS-ESS3 | applied | 34 | 0.96 | 0.06 | 0.82 | 0.00 |
| HS-ESS3 | conceptual | 32 | 0.99 | 0.03 | 0.91 | 0.00 |
| HS-ESS3 | recall | 32 | 0.99 | 0.03 | 0.91 | 0.00 |
| HS-ETS1 | applied | 34 | 0.96 | 0.09 | 0.77 | 0.00 |
| HS-ETS1 | conceptual | 32 | 0.99 | 0.03 | 0.91 | 0.00 |
| HS-ETS1 | recall | 31 | 1.00 | 0.00 | 1.00 | 0.00 |

### 4.3 Discussion

Our results highlight that you do not need large, closed-weight LLMs to have a model capable of understanding secondary science topics aligned with the NGSS. In our results, OSS-20b and Nemotron 3 Nano 30b a3 scored quite well across all metrics, outperforming models significantly larger than themselves. Of interest is that the strongly performing models generally performed well across all areas of the NGSS, and across all item types.

An important question arises from our work: is our benchmark too easy? According to classical test theory (see De Champlain (2010) for a brief primer), this may be the case. If we examine the results in Table 5, we see that a number of questions are considered trivial, and item difficulty was generally high. Meanwhile, item discrimination was low, and many models scored well on the benchmark itself. A research question for future investigation is if this is because the questions are too easy and could be improved, or if models are simply at a point where secondary science is relatively easy for them. Since this is a fully synthetic benchmark, there is ample room for human-curated item comparisons in future research. We approach this question in Part III.

Overall, from our benchmarking results we can conclude two things: model size did not correspond to higher scores in secondary science and model choice matters. This work shows that model choice in educational contexts should be an intentional decision-making process. Moreover, among calls for privacy-protecting, local deployment of LLMs in schools (Schroeder et al., 2026), this work provides evidence that small, locally deployed, open-weight models can provide competitive knowledge of scientific content aligned with the NGSS, and this is not restricted to recall type questions.

## 5. Part III

A key limitation of the aforementioned work was that the items within the benchmark were synthetically generated, without any human oversight. Given the strong scores produced by the models in Part II, Part III involved having a human review a subset of the question and answer pairs to see if they held validity with the respective standards. Specifically, we address the question, RQ3: To what extent does a former K-12 science teacher feel the items align with the NGSS standards?

### 5.1 Methods

*Dataset.* The dataset for human review consisted of 120 question and answer pairs from the middle school and high school datasets, for a total of 240 question and answer pairs. The questions were evenly distributed across areas of the NGSS and relatively evenly distributed by question type. Table A1 shows the distribution of questions by area and question type.

*Human Reviewer.* Author YA served as a K-12 teacher for 10 years, specifically teaching science courses for grades 7-12 (middle school and high school) before returning to academia to pursue graduate degrees. YA had no role in creating the synthetically generated items, so her review of the items was her first time seeing them. She was specifically asked to evaluate 1) if the question-answer pair aligned with the respective standard, and 2) if the answer provided was in fact correct.

### 5.2 Results

Overall, 236 of the 240 reviewed items (98.3%) were judged to align with their intended NGSS standards, while 4 items (1.7%) were identified as misaligned. No incorrect reference answers were identified during the review. Table 7 summarizes the four items that were judged not to align with their assigned standards and provides the reviewer's rationale.

For the middle school dataset, 119 of 120 items (99.2%) were judged to align with their assigned standards. One item was identified as misaligned. For the high school dataset, 177 of 120 items (97.5%) were judged to align with their assigned standards, with three items identified as misaligned. Across the four rejected items, the questions generally addressed content related to the target standard but did not fully reflect critical components of the NGSS performance expectation. Misalignments most commonly occurred when a question focused on a related concept while omitting the primary relationship, scientific practice, or representation emphasized by the standard.

YA also noted several characteristics of the generated items. None of the reviewed questions included graphical elements such as diagrams, figures, or graphs. Additionally, data interpretation tasks were absent from the reviewed sample. Standards involving mathematical reasoning often expressed relationships through text descriptions rather than through formulas, symbols, or quantitative representations. These observations suggest that while the generated items are generally aligned with NGSS content expectations, some scientific practices are underrepresented in the benchmark.

During the review process, YA also observed that several NGSS standards were not represented in the sampled items. For middle school standards, the review sample did not include items associated with MS-PS1-6, MS-PS2-5, MS-PS3-5, MS-LS2-5, MS-LS1-5, MS-LS4-6,

MS-ESSS2-5, MS-ESSS2-6, and MS-ESSS3-5. For high school standards, the review sample did not include items associated with HS-PS1-5 through HS-PS1-8, HS-PS2-5, HS-PS2-6, HS-PS3-5, HS-PS4-5, HS-ESSS1-5, HS-ESSS1-6, HS-LS1-5 through HS-LS1-7, HS-LS2-5 through HS-LS2-8, HS-LS4-5, HS-LS4-6, and HS-ESSS2-5 through HS-ESSS2-7. As Part III examined only a subset of the benchmark, these observations should not be interpreted as evidence that the standards were absent from the full dataset.

**Table 6.** Misaligned Benchmark Items Identified Through Human Review

| **Middle School** | | | | | | |
|---|---|---|---|---|---|---|
| **standard** | **NGSS Standards** | **Question type** | **Question** | **Reference Answer** | **Human Review** | **Human Rationale for 0** |
| MS-PS4-1 | “Use mathematical representations to describe a simple model for waves that includes how the amplitude of a wave is related to the energy in a wave.” (Achieve, Inc., 2013) | recall | What is the unit used to measure frequency? | Frequency is measured in hertz (Hz), where one hertz equals one wave cycle per second. | 0 | not aligned, the standard mentioned about the relation between amplitude and energy so the question is supposed to ask more about amplitude, instead of frequency |
| **High School** | | | | | | |
| HS-ESS3-3 | “Create a computational simulation to illustrate the relationships among the management of natural resources, the sustainability of human populations, and biodiversity.” (Achieve, Inc., 2013) | recall | What is an ecosystem service? | An ecosystem service is a benefit that humans receive from natural ecosystems, such as clean water filtration, pollination of crops, carbon storage, and flood control. | 0 | Rejected because the question asks for the definition of an ecosystem service rather than assessing understanding of the relationships among natural resource management, sustainability, and biodiversity that are the focus of the standard. |
| HS-LS1-3 | “Plan and conduct an investigation to provide evidence that | recall | What is the difference between negative feedback | Negative feedback reverses or counteracts a change to bring a system back to its | 0 | Rejected because the question focuses on types of feedback rather |
| | feedback mechanisms maintain homeostasis." (Achieve, Inc., 2013) | | and positive feedback? | set point, while positive feedback amplifies or intensifies a change, moving the system further from its original state. | | than their role in maintaining homeostasis, which is the primary focus of the standard. |
| HS-LS2-2 | "Use mathematical representations to support and revise explanations based on evidence about factors affecting biodiversity and populations in ecosystems of different scales." (Achieve, Inc., 2013) | conceptual | How does the introduction of an invasive species cause native population sizes to decline? | Invasive species compete with native species for the same limited resources—such as food, space, or light—and often lack natural predators in the new environment, giving them a competitive advantage that reduces native population sizes. | 0 | no mathematical representation in the question |

### 5.3 Discussion

The Part III findings provide evidence towards the validity of the benchmark built in Parts I and II. Of the 240 reviewed question-answer pairs, 236 (98.3%) aligned with their intended NGSS standards, a result that points to a synthetic data generation pipeline capable of producing items consistent with NGSS content expectations. However, the small number of misaligned items points to real limitations in synthetic benchmark generation. In each case, the generated question stayed close to the target standard but missed a key piece of the performance expectation. This pattern suggests that synthetic items can lean too heavily on associated content knowledge while leaving out the scientific practices or conceptual relationships the standard is actually built around. High benchmark coverage does not necessarily guarantee complete alignment with the intended learning objectives.

The review also surfaced some structural gaps worth noting for future works. Graphical representations, data interpretation tasks, and mathematical representations were largely missing from the sample. This is notable because some NGSS performance standards require students to engage with evidence, analyze data, interpret visual information, or use mathematical reasoning. Although text-based questions can assess many aspects of science knowledge, future iterations of the benchmark could incorporate a wider range of item formats to better reflect the breadth of practices emphasized within the NGSS.

## 6. General Discussion

Our work demonstrates a replicable pipeline for creating standards-anchored benchmarks for LLMs in educational contexts. As with any benchmark, this initial exploration of model performance should be interpreted as *indications of model performance* in specific domains rather than concrete, broadly generalizable claims about model *competence*. As such, we position the remainder of our discussion as the limitations and implications for future research and implications for practice.

### *6.1 Limitations and Implications for Research*

It is critically important that our results are not overgeneralized. The work presented here holds value as a starting point for benchmarking LLMs knowledge of secondary NGSS-aligned content knowledge. We argue that work such as this is not the ending point, but rather a reference from which further experimentation and comparisons can be made.

Although Part III provided evidence supporting the validity of the benchmark, only a sample of benchmark items underwent human review. Future studies could expand this validation effort by increasing the number of expert reviewers, reviewing a larger proportion of benchmark items, and conducting additional forms of validity analysis. Item quality is of utmost importance in any benchmark, and we demonstrated both how to develop such a benchmark with quality-assurance gates, as well as evaluate a variety of initial metrics for model performance across said benchmarks. Specifically, we believe that this work presents an ideal environment for gaining initial insights into model performance across a broad range of secondary school science areas, as well as an ideal testbed for testing custom models, model harnesses (which often provide models with a series of tools they can use), or multi-agent systems. This work also provides a foundation for those fine-tuning or using other post-training methods (like reinforcement learning) to develop models specifically for K-12 educational contexts and content areas. A sample use case would be trying to improve the performance of models small enough to run in web-browers on the individual user's machine. Many such models currently exist (e.g., Amini et al., 2025), as does the technology to run them in-browser (e.g., Chen et al., 2025), and this represents one possible way to preserve learner privacy in school settings (Schroeder et al., 2026). With this work as the foundation, a team could benchmark a tiny LLM's performance across scientific areas, then use supervised fine-tuning (Pareja et al., 2025), reinforcement learning (e.g., Yu et al., 2026), or a pipeline with a combination of these features (e.g., OLMo et al., 2024) to improve model performance, while maintaining a baseline to compare against (i.e., these benchmarks). Such work is impossible without benchmarks such as those created here, and these benchmarks were in fact created, in part, to enable our team to test post-training pipelines for models intended to be used in educational settings.

As noted, one possible limitation of this work is that the item-level metrics were generally poor, but we suspect this was because models scored so well rather than innate problems with the items themselves. We believe this to be the case due to our human review in Part III, where we did not see extensive issues with questions, but future work could include a detailed human review of every item generated. Evidence around using LLMs for item generation show that LLMs are capable of creating items with strong psychometric properties (Crk, & Gultepe, 2026; Wróblewska et al. 2025), so meaningful next steps would include having a human-in-the-loop procedure (Dellerman et al., 2019) where humans review and edit the items generated by the LLMs, or even fully human created benchmarks. Human curation may be important during the item review phase specifically, as previous research found that humans pick up on things LLMs may not when reviewing test items (Kim et al., 2025), a trend we saw in a small number of items in our human-reviewed sample in Part III. These types of human curation could lend insights into if human involvement improves item-level metrics, or if the models really understand secondary science as well as our benchmarks imply. Research is needed to understand if there is a meaningful value-add from having humans in the loop with content that should, in concept, be understood by many LLMs given that many score well on more complex scientific benchmarks such as GPQA (Rein et al., 2023).

A final future extension of this work where research is direly needed is understanding the intersection of model content knowledge and model pedagogical knowledge. As noted earlier, research has examined both individually, but research at the intersection is less common. An example of the intersection of these areas of work include examining how well LLMs can generate appropriate test items in specific domains (Chen et al., 2024; Jaldi et al., 2026; Shirkar et al., 2025). However, these types of benchmarks lack measures of pedagogical understanding in other contexts such as providing feedback to students in a multi-turn conversation. Being able to meaningfully measure and evaluate multi-turn dialogue with students, ensuring alignment with evidence-based feedback and scaffolding strategies, while maintaining accurate content knowledge, could represent a gold standard in benchmarking LLMs for education.

### *6.2 Implications for Practice*

Educators who are new to locally-deployed, open-weight LLMs may struggle with model selection due to the sheer number of models available. Our work provides concrete evidence that some small, open-weight LLMs can convey meaningful NGSS-aligned secondary school content knowledge. Specifically, our results showed that both OSS-20b and Nemotron 3 Nano 30b a3 may offer plausible options for educators, among others, although we note that there are other considerations that are outside the scope of this benchmark (e.g., model guardrails) that should also be investigated. We hope to see others use our benchmark with a wider variety of models to create a more comprehensive database of model performance. Moreover, for those building and refining LLMs for use in education, our benchmarks represent an evaluation metric from which to compare base model performance and training improvement.

## 7. Conclusion

LLMs are being broadly implemented in educational settings, but the evidence supporting the choice of any specific model has been sparse or non-existent. In an effort to bridge this cap, we developed two NGSS-aligned benchmarks for secondary science. Our results show that a number of small open-weight LLMs that can be easily deployed on consumer hardware have strong performance across both middle school and high school level NGSS areas. Moreover, our pipeline creates a foundation for further testing and development of specialized models. We encourage researchers to pursue the development of additional benchmarks that bridge the gap between content knowledge and pedagogical knowledge, particularly across multi-turn conversations.

**Declaration of generative AI and AI-assisted technologies in the manuscript preparation process**
During the preparation of this work, the author(s) used Whisper-base to transcribe the first author's dictated notes and Gemma 4 24ba-4b to convert these notes into an initial draft of some paper sections. All citations were added by the authors. After using this tool, the author(s) reviewed and edited the content and take(s) full responsibility for the content of the published article.

**Funding**: This material is based upon work supported by the National Science Foundation and the Institute of Education Sciences under Grant DRL-2229612, and the U.S. Department of Education under award P116J250669. Any opinions, findings, and conclusions or recommendations expressed in this material are those of the author(s) and do not necessarily reflect the views of the National Science Foundation or the U.S. Department of Education.

Appendix A

**Table A1.** The distribution of questions reviewed by the human expert, by question area and question type.

**Middle School**

| Area | Standard | Recall | Conceptual | Applied | Total |
|---|---|---|---|---|---|
| MS-ESS1 | MS-ESS1-1 | 1 | 1 | 1 | 3 |
| | MS-ESS1-2 | 1 | 1 | 1 | 3 |
| | MS-ESS1-3 | 1 | 1 | 1 | 3 |
| | MS-ESS1-4 | 0 | 0 | 1 | 1 |
| MS-ESS2 | MS-ESS2-1 | 1 | 1 | 1 | 3 |
| | MS-ESS2-2 | 1 | 1 | 1 | 3 |
| | MS-ESS2-3 | 1 | 1 | 1 | 3 |
| | MS-ESS2-4 | 0 | 0 | 1 | 1 |
| MS-ESS3 | MS-ESS3-1 | 1 | 1 | 1 | 3 |
| | MS-ESS3-2 | 1 | 1 | 1 | 3 |
| | MS-ESS3-3 | 1 | 1 | 1 | 3 |
| | MS-ESS3-4 | 0 | 0 | 1 | 1 |
| MS-ETS1 | MS-ETS1-1 | 1 | 1 | 1 | 3 |
| | MS-ETS1-2 | 1 | 1 | 1 | 3 |
| | MS-ETS1-3 | 1 | 1 | 1 | 3 |
| | MS-ETS1-4 | 0 | 0 | 1 | 1 |
| MS-LS1 | MS-LS1-1 | 1 | 1 | 1 | 3 |
| | MS-LS1-2 | 1 | 1 | 1 | 3 |
| | MS-LS1-3 | 1 | 0 | 1 | 2 |
| | MS-LS1-4 | 1 | 0 | 1 | 2 |
| MS-LS2 | MS-LS2-1 | 1 | 1 | 1 | 3 |
| | MS-LS2-2 | 1 | 1 | 1 | 3 |
| | MS-LS2-3 | 1 | 1 | 1 | 3 |
| | MS-LS2-4 | 0 | 0 | 1 | 1 |
| MS-LS3 | MS-LS3-1 | 2 | 2 | 2 | 6 |
| | MS-LS3-2 | 1 | 1 | 2 | 4 |
| MS-LS4 | MS-LS4-1 | 1 | 2 | 1 | 4 |
| | MS-LS4-2 | 1 | 1 | 1 | 2 |
| | MS-LS4-3 | 1 | 0 | 1 | 2 |
| | MS-LS4-4 | 0 | 0 | 1 | 1 |
| MS-PS1 | MS-PS1-1 | 1 | 1 | 1 | 3 |
| | MS-PS1-2 | 1 | 0 | 1 | 2 |
| | MS-PS1-3 | 1 | 0 | 1 | 2 |
| | MS-PS1-4 | 0 | 1 | 1 | 2 |
| | MS-PS1-5 | 0 | 1 | 0 | 1 |
| MS-PS2 | MS-PS2-1 | 1 | 1 | 1 | 3 |
| | MS-PS2-2 | 1 | 1 | 1 | 3 |
| | MS-PS2-3 | 1 | 1 | 1 | 3 |
| | MS-PS2-4 | 0 | 0 | 1 | 1 |
| MS-PS3 | MS-PS3-1 | 1 | 1 | 1 | 3 |
| | MS-PS3-2 | 1 | 1 | 1 | 3 |
| | MS-PS3-3 | 1 | 1 | 1 | 3 |
| | MS-PS3-4 | 0 | 0 | 1 | 1 |
| MS-PS4 | MS-PS4-1 | 1 | 1 | 2 | 4 |
| | MS-PS4-2 | 1 | 1 | 1 | 3 |
| | MS-PS4-3 | 1 | 1 | 1 | 3 |

**High School**

| Area | Standard | Recall | Conceptual | Applied | Total |
|---|---|---|---|---|---|
| HS-ESS1 | HS-ESS1-1 | 1 | 1 | 1 | 3 |
| | HS-ESS1-2 | 1 | 1 | 1 | 3 |
| | HS-ESS1-3 | 1 | 1 | 1 | 3 |
| | HS-ESS1-4 | 0 | 0 | 1 | 1 |
| HS-ESS2 | HS-ESS2-1 | 1 | 1 | 1 | 3 |
| | HS-ESS2-2 | 1 | 1 | 1 | 3 |
| | HS-ESS2-3 | 1 | 1 | 1 | 3 |
| | HS-ESS2-4 | 0 | 0 | 1 | 1 |
| HS-ESS3 | HS-ESS3-1 | 1 | 1 | 1 | 3 |
| | HS-ESS3-2 | 1 | 1 | 1 | 3 |
| | HS-ESS3-3 | 1 | 1 | 1 | 3 |
| | HS-ESS3-4 | 0 | 0 | 1 | 1 |
| HS-ETS1 | HS-ETS1-1 | 1 | 1 | 1 | 3 |
| | HS-ETS1-2 | 1 | 1 | 1 | 3 |
| | HS-ETS1-3 | 1 | 1 | 1 | 3 |
| | HS-ETS1-4 | 0 | 0 | 1 | 1 |
| HS-LS1 | HS-LS1-1 | 1 | 1 | 1 | 3 |
| | HS-LS1-2 | 1 | 1 | 1 | 3 |
| | HS-LS1-3 | 1 | 1 | 1 | 3 |
| | HS-LS1-4 | 0 | 0 | 1 | 1 |
| HS-LS2 | HS-LS2-1 | 1 | 1 | 1 | 3 |
| | HS-LS2-2 | 1 | 1 | 1 | 3 |
| | HS-LS2-3 | 1 | 1 | 1 | 3 |
| | HS-LS2-4 | 0 | 0 | 1 | 1 |
| HS-LS3 | HS-LS3-1 | 1 | 1 | 2 | 4 |
| | HS-LS3-2 | 1 | 1 | 1 | 3 |
| | HS-LS3-3 | 1 | 1 | 1 | 3 |
| HS-LS4 | HS-LS4-1 | 1 | 0 | 1 | 2 |
| | HS-LS4-2 | 1 | 1 | 1 | 3 |
| | HS-LS4-3 | 1 | 1 | 1 | 3 |
| | HS-LS4-4 | 0 | 1 | 1 | 2 |
| HS-PS1 | HS-PS1-1 | 1 | 1 | 1 | 3 |
| | HS-PS1-2 | 1 | 1 | 1 | 3 |
| | HS-PS1-3 | 1 | 1 | 1 | 3 |
| | HS-PS1-4 | 0 | 0 | 1 | 1 |
| HS-PS2 | HS-PS2-1 | 1 | 1 | 1 | 3 |
| | HS-PS2-2 | 1 | 1 | 1 | 3 |
| | HS-PS2-3 | 1 | 1 | 1 | 3 |
| | HS-PS2-4 | 0 | 0 | 1 | 1 |
| HS-PS3 | HS-PS3-1 | 1 | 1 | 1 | 3 |
| | HS-PS3-2 | 1 | 1 | 1 | 3 |
| | HS-PS3-3 | 1 | 1 | 1 | 3 |
| | HS-PS3-4 | 0 | 0 | 1 | 1 |
| HS-PS4 | HS-PS4-1 | 1 | 1 | 1 | 3 |
| | HS-PS4-2 | 1 | 1 | 1 | 3 |
| | HS-PS4-3 | 1 | 1 | 1 | 3 |
| | HS-PS4-4 | 0 | 0 | 1 | 1 |